\pdfoutput=1

\documentclass[11pt]{article}

\usepackage[final]{acl}
\usepackage{times}
\usepackage{latexsym}
\usepackage[table]{xcolor}
\usepackage[T1]{fontenc}
\usepackage[utf8]{inputenc}
\usepackage{microtype}
\usepackage{inconsolata}
\usepackage{graphicx}
\usepackage{amsmath}
\usepackage{amsthm}
\usepackage{csquotes}
\usepackage{amssymb}
\usepackage{algorithm}
\usepackage{booktabs}  
\usepackage{algorithmic}

\usepackage{multirow} % 支持跨行
\usepackage{array}

\usepackage{subcaption}

\usepackage{enumitem}

\makeatletter
\newenvironment{breakablealgorithm}
  {% \begin{breakablealgorithm}
   \begin{center}
     \refstepcounter{algorithm}% New algorithm
     \hrule height.8pt depth0pt \kern2pt% \@fs@pre for \@fs@ruled
     \renewcommand{\caption}[2][\relax]{% Make a new \caption
       {\raggedright\textbf{\ALG@name~\thealgorithm} ##2\par}%
       \ifx\relax##1\relax % #1 is \relax
         \addcontentsline{loa}{algorithm}{\protect\numberline{\thealgorithm}##2}%
       \else % #1 is not \relax
         \addcontentsline{loa}{algorithm}{\protect\numberline{\thealgorithm}##1}%
       \fi
       \kern2pt\hrule\kern2pt
     }
  }{% \end{breakablealgorithm}
     \kern2pt\hrule\relax% \@fs@post for \@fs@ruled
   \end{center}
  }
\makeatother

\title{TCPO: Thought-Centric Preference Optimization for\\ Effective Embodied Decision-making}

\author{Kechen Jiao$^{1\ 2}$\footnotemark[1]\footnotemark[3], Zhirui Fang$^{1}$\footnotemark[1], Jiahao Liu$^{2}$, Bei Li$^{2}$, Qifan Wang$^{5}$, Xinyu Liu$^{3}$,\\ \textbf{Junhao Ruan$^{3}$, Zhongjian Qiao$^{1}$, Yifan Zhu$^{4}$, Yaxin Xu$^{6}$, Jingang Wang$^{2}$, Xiu Li$^{1}$\footnotemark[2]}\\
$^{1}$Tsinghua University, $^{2}$Meituan
$^{3}$Northeastern University \\
$^{4}$Beijing University of Posts and Telecommunications\\
$^{5}$Meta AI, 
$^{6}$Wuhan University}

\begin{document}
\maketitle

\renewcommand{\thefootnote}{\fnsymbol{footnote}}
\footnotetext[1]{Equal Contribution}
\footnotetext[2]{Corresponding author}
\footnotetext[3]{Work performed while an intern at Meituan.}
\begin{abstract}

Using effective generalization capabilities of vision language models (VLMs) in context-specific dynamic tasks for embodied artificial intelligence remains a significant challenge. Although supervised fine-tuned models can better align with the real physical world, they still exhibit sluggish responses and hallucination issues in dynamically changing environments, necessitating further alignment. Existing post-SFT methods, reliant on reinforcement learning and chain-of-thought (CoT) approaches, are constrained by sparse rewards and action-only optimization, resulting in low sample efficiency, poor consistency, and model degradation.
To address these issues, this paper proposes Thought-Centric Preference Optimization (TCPO) for effective embodied decision-making. Specifically, TCPO introduces a stepwise preference-based optimization approach, transforming sparse reward signals into richer step sample pairs. It emphasizes the alignment of the model's intermediate reasoning process, mitigating the problem of model degradation. Moreover, by incorporating Action Policy Consistency Constraint (APC), it further imposes consistency constraints on the model output. 
Experiments in the ALFWorld environment demonstrate an average success rate of \textbf{26.67\%}, achieving a \textbf{6\%} improvement over RL4VLM and validating the effectiveness of our approach in mitigating model degradation after fine-tuning. These results highlight the potential of integrating preference-based learning techniques with CoT processes to enhance the decision-making capabilities of vision-language models in embodied agents.
\end{abstract}

\section{Introduction}

Large Language Models (LLMs) and Large Multi-modal Models (LMMs) have demonstrated exceptional capabilities in natural language understanding and generation \citep{brown2020language, achiam2023gpt}. Recent advances extend their applications to managing AI models for complex multi-modal tasks \citep{shen2024hugginggpt, lu2024chameleon}, mastering strategic games like TextWorld \citep{yao2022react}, Handi \citep{hu2023language}, and Minecraft \citep{wang2023voyager}, as well as enabling robotic interactions through physical deployments \citep{ahn2022can, driess2023palm, ahn2022can}.

\begin{figure}[t]
    \centering
    \includegraphics[width=1\linewidth]{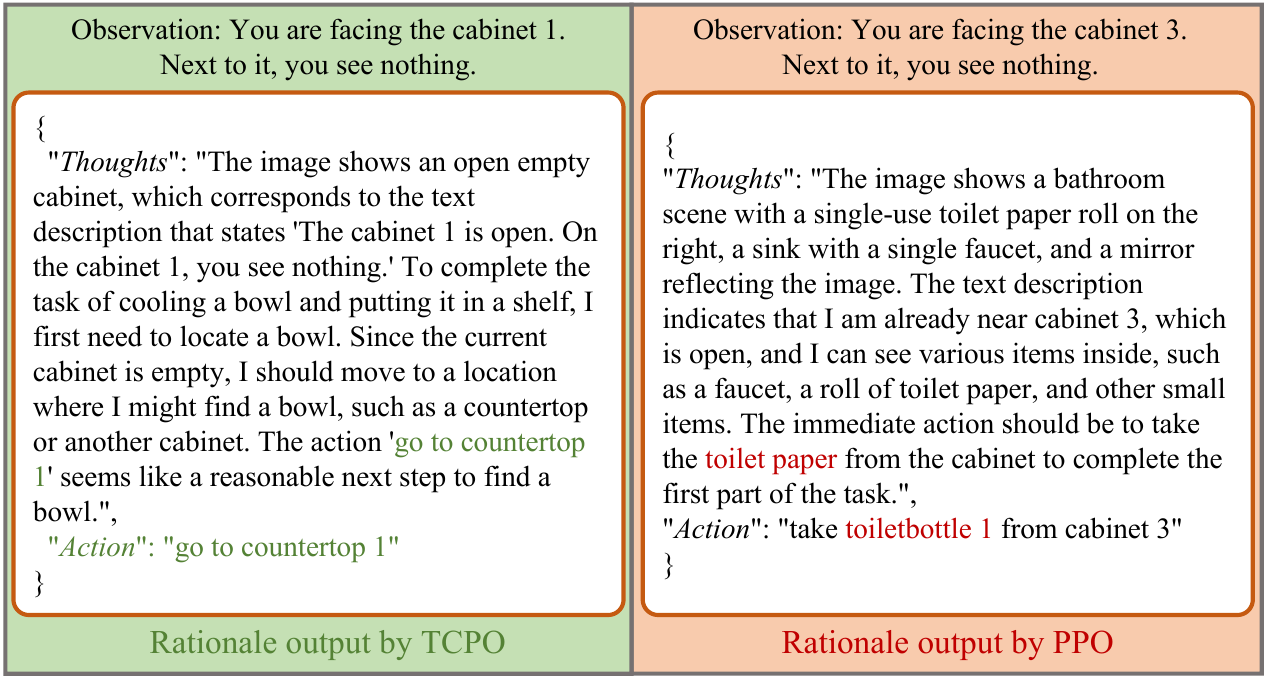}
    \caption{Comparison results of our TCPO and PPO methods. In our TCPO method, we emphasize the logical consistency of actions generated by rationale and incorporate an Action Probability Consistency constraint (APC). In contrast, traditional PPO methods may compromise consistency during training, resulting in the generation of illegal actions, as shown on the right.}
    \label{fig:kl-cases}
    \vspace{-5mm}
\end{figure}
Embodied AI research has predominantly concentrated on developing foundational models to augment semantic comprehension and operational capacities of LLMs and multi-modal systems through robotic sensory inputs \citep{mu2023embodiedgptvisionlanguagepretrainingembodied, kim2024openvlaopensourcevisionlanguageactionmodel, xu2024surveyroboticsfoundationmodels}. These initiatives aim to bridge the disconnect between pre-trained models' knowledge and physical environments, typically employing supervised fine-tuning (SFT) or LoRA adaption techniques \citep{hu2021lora} to improve visual understanding, planning proficiency, and action strategy generation from multi-modal inputs. However, such static adaptation approaches prove inadequate for dynamic environments, prompting development of two post-SFT enhancement strategies: dynamic replanning and reinforcement learning integration.

Replanning methodologies address environmental dynamics through chain-of-thought reasoning and task decomposition \citep{mu2023embodiedgptvisionlanguagepretrainingembodied, song2023llmplannerfewshotgroundedplanning}, enabling real-time plan updates when environmental states change. This approach introduces adaptive error correction and contingency handling to static planning frameworks. Reinforcement learning extensions further align models with dynamic requirements through various implementations. LLaRP \citep{szot2024largelanguagemodelsgeneralizable} integrates policy heads into language models, RL4VLM \citep{zhai2024fine} employs Proximal Policy Optimization (PPO) \citep{schulman2017proximalpolicyoptimizationalgorithms} for decision-making, and TWOSOME \citep{tan2024trueknowledgecomespractice} aligns action probabilities using reinforcement principles \citep{sutton2018reinforcement}.
Despite these advancements, practical deployment faces two critical challenges: 1) \textit{the prevalence of sparse environmental rewards that escalate exploration costs}, and 2) \textit{the inherent conflict between reinforcement optimization and linguistic consistency preservation}. As demonstrated in Figure \ref{fig:kl-cases}, conventional reinforcement paradigms that optimize action probabilities or joint thought-action alignment tend to disrupt internal linguistic coherence of models, ultimately degrading response quality despite improved environmental adaptation.

To address these challenges, we propose that optimization should focus on enhancing the quality of Chain-of-Thought (CoT) reasoning rather than final actions, as strategic decisions inherently emerge from this cognitive process. Our solution leverages a step-wise Direct Preference Optimization (DPO) framework to maximize sample efficiency. Unlike conventional reinforcement learning requiring dense rewards, preference learning effectively utilizes entire trajectories – including zero-return samples through negative pair construction – demonstrating enhanced learning capacity for sparse-reward scenarios and long-horizon tasks.
We introduce \textbf{Thought-Centric Preference Optimization (TCPO)}, a paradigm prioritizing rationale refinement over action selection to address composite error propagation in multi-step reasoning while strengthening step-wise determinism (see Section~\ref{sec:ourDPO}). Experiments verify the superior capability of TCPO in learning deterministic strategies and resolving credit assignment challenges. Furthermore, we establish the \textbf{Action Policy Consistency Constraint (APC)} to preserve the model's intrinsic consistency, ensuring actions strictly derive from CoT processes through constrained policy optimization. 
Our main contributions can be summarized as following:

\begin{itemize}
    \item We present \textbf{TCPO}, an algorithmic framework employing stepwise alignment methodology to coordinate the CoT process in embodied agents via environmental interactions. The framework strengthens model coherence through strategic determinism optimization while maintaining online adaptability.

    \item We introduce the novel Action Policy Consistency Constraint (APC), enforcing alignment with the pre-trained model's action conditional distributions to address policy consistency deterioration during online adaptation.

    \item Our experimental evaluation on GymCards and ALFWorld demonstrate that the proposed approach achieves a \textbf{6\%} improvement in average task success rate compared to state-of-the-art RL4VLM baselines.
\end{itemize}

\section{Related Work}

\paragraph{Embodied Agent with LLMs}
Recent works highlight the importance of LLMs in interaction and decision-making~\citep{abramson2020imitating, karamcheti2022lila, li2022pre}, and their application in robot navigation~\citep{parisi2022unsurprising, hong2021vln, majumdar2020improving} and manipulation~\citep{jiang2022vima, ren2023leveraging, karamcheti2022lila}. A growing body of research leverages LLMs to enhance planning and reasoning in embodied agents. SayCan~\citep{ahn2022can} combines LLM probabilities with a value function to assess candidate actions. \citet{zeng2022socratic} integrate LLMs with visual-language models and pre-trained language-conditioned policies~\citep{shridhar2022cliport} for open vocabulary tasks. \citet{huang2022language} show that LLMs can plan and execute household tasks by grounding actions to a predefined list. Inner Monologue~\citep{huang2022inner} extends SayCan with a closed-loop principle, also applied in works like~\citep{yao2023react, huang2022inner, kim2024language, singh2023progprompt, liang2023code, shinn2023reflexionlanguageagentsverbal, wang2023describe} to refine plans based on environment feedback for tasks such as automation and Minecraft. Approaches like~\citep{zheng2023synapse} use LLMs to generate temporal-abstracted actions, while \citet{dasgupta2023collaborating} employ LLMs for planning and success detection in RL-trained agents. While these methods show strong results, they depend heavily on powerful LLMs like GPT-4 and PaLM~\citep{chowdhery2023palm}, which may not be suitable for smaller models like LLaMA-7B with weaker reasoning abilities.

Similarly, GLAM~\citep{carta2023grounding} uses RL finetuning for grounding LLMs but focuses on simple actions (e.g., turn left, go forward) in toy environments like BabyAI~\citep{chevalier2018babyai}, using a much smaller LLM (Flan-T5-780M). These simple actions, with fewer tokens and less meaningful semantics, underutilize LLM capabilities and fail to address prompt design issues and action space imbalance, leading to instability and poor robustness.

\paragraph{Preference Learning}
Preference learning has become a key area in machine learning, focusing on developing models that capture human preferences from observational data. 
Current preference learning methods are typically categorized into pointwise, pairwise, and listwise approaches. Among these, Direct Preference Optimization (DPO)~\citep{rafailov2024directpreferenceoptimizationlanguage} has emerged as a novel approach, directly optimizing user preferences without intermediary ranking steps. This method enhances alignment with user preferences by constructing loss functions that reflect them directly. \citet{chen2024optuneefficientonlinepreference} introduces OPTune, an efficient online preference tuning method in RLHF. OPTune improves training speed and model alignment by selectively regenerating low-reward responses and focusing on response pairs with larger reward gaps using a weighted DPO loss.

Recent studies have expanded DPO's applications. Step-DPO~\citep{lai2024stepdpostepwisepreferenceoptimization} enhances DPO for tasks requiring long-chain reasoning, like mathematical problem-solving, by optimizing individual reasoning steps and improving both factuality and reasoning in large language models. \citet{pal2024smaugfixingfailuremodes} advanced DPO’s practical applications in sentiment-aware recommendations through DPO-Positive, which integrates sentiment information into the recommendation process, leading to more accurate and user-aligned outcomes.

\section{Methodology}
\label{sec:ourDPO}

\begin{figure*}
    \centering
    \includegraphics[width=0.95\linewidth]{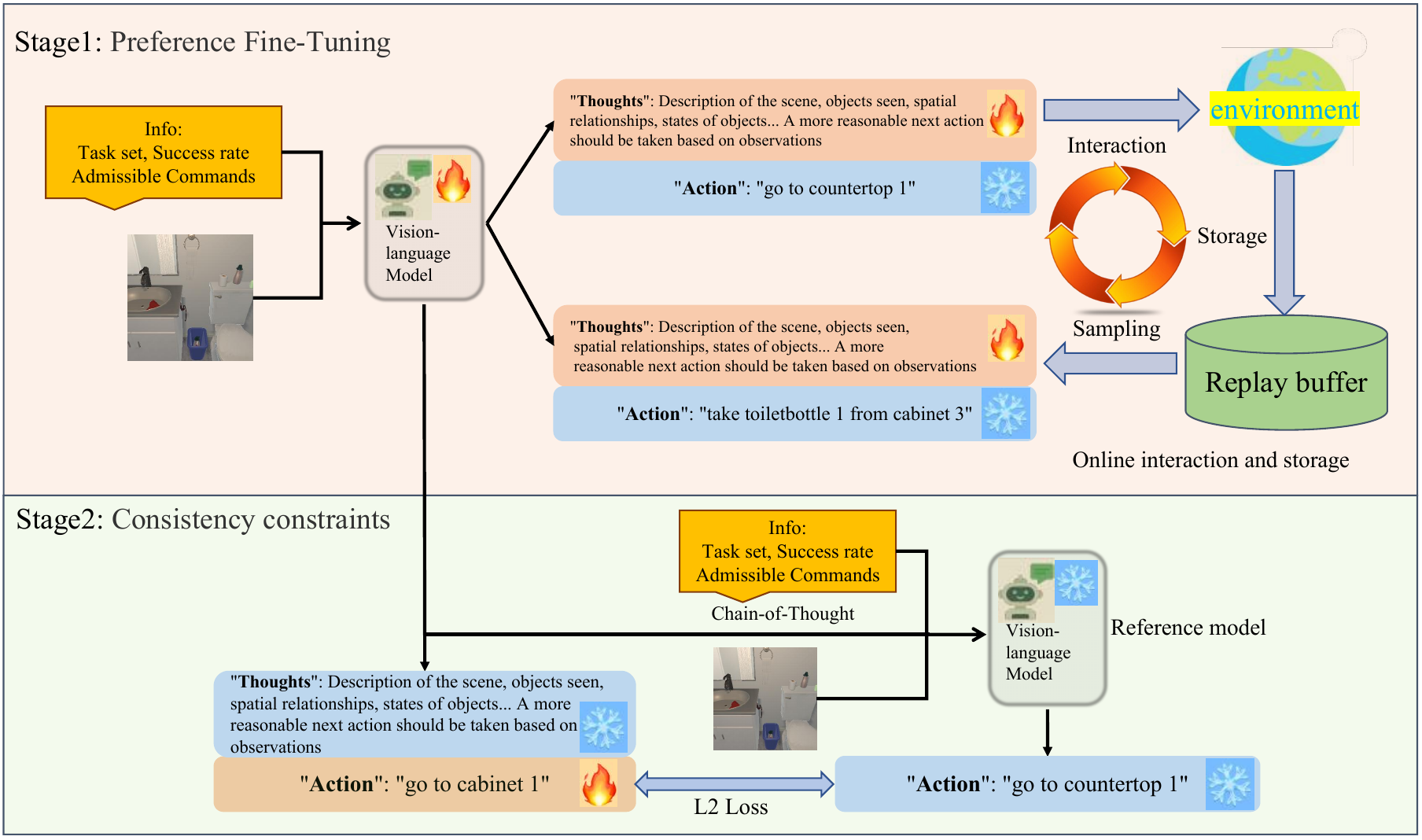}
    \caption{Overview of TCPO framework. The upper stage implements preference-driven CoT fine-tuning: The VLM processes environmental observations through CoT reasoning, generating spatial analyses and executable actions. Online interaction stores decision trajectories in a replay buffer, while contrastive learning with step-wise preference judgments optimizes thought-action distributions. The lower stage enforces APC through L2 loss regularization. This preserves pretrained thought-action mappings while constraining outputs to valid operations, as evidenced by comparative case studies. The  `flame' and `snowflake' symbols indicate whether gradient backpropagation is applied to corresponding parameters during that training stage.}
    \label{fig:main_pic}
    \vspace{-4mm}
\end{figure*}

Our proposed Thought-Centric Preference Optimization (TCPO) framework employs a replanning-enabled algorithmic architecture to ensure robust adaptability in dynamic environments, comprising two core components.
The \textit{Preference-Aware Fine-Tuning} component introduces a stepwise preference learning mechanism that reframes the alignment task as a cross-entropy-guided classification problem, allowing for dense preference supervision and more efficient policy optimization.
Sample efficiency is further improved through trajectory repurposing, where typically discarded zero-return trajectories are leveraged to generate auxiliary training pairs via contrastive sampling.
The \textit{Action Policy Consistency Constraints} component enforces coherence between intermediate reasoning states and final action outputs, effectively mitigating model degradation observed in conventional chain-of-thought approaches.
We present the two components in the following subsections.

\subsection{Sample Pairs Construction}

Unlike traditional MLP-based policy networks restricted to predefined action spaces, VLM policies exhibit unique advantages through their natural language generation capabilities. This enables explicit CoT reasoning that facilitates systematic environment exploration via intermediate rationalization steps preceding final action selection. However, RL-based fine-tuning of VLM policies $\pi_\theta$ introduces critical challenges arising from sparse reward signals. Specifically, the episodic nature of embodied interactions yields predominantly non-informative state transitions where reward feedback $r_t=0$ for most timesteps $t$. Under standard PPO frameworks, these zero-reward transitions provide negligible gradient signals due to their dependence on advantage estimation, resulting in suboptimal sample efficiency during policy adaptation. While conventional approaches often resort to manual reward shaping to mitigate sparsity, our method addresses this through contrastive trajectory pair construction.
For any timestep $t$, we generate preference tuples:
\begin{equation}
\small
\begin{aligned}
\mathcal{P}_t &= \left\langle \tau_{\mathrm{win}}^t, \tau_{\mathrm{lose}}^t \right\rangle \\
&= \left\langle \left\{a_t^{(1)}, r_t^{(1)}, \tau_{1:t-1}\right\}, \left\{a_t^{(2)}, r_t^{(2)}, \tau_{1:t-1}\right\} \right\rangle
\end{aligned}
\end{equation}
where $\tau_{\mathrm{win}}^t$ denotes the preferred trajectory segment with comparatively higher reward $r_t^{(1)} > r_t^{(2)}$. This construction transforms sparse scalar rewards into relative preference rankings across trajectory segments, enabling three key advancements: (1) Effective utilization of suboptimal transitions where $r_t^{(i)} > 0$ but $r_t^{(i)} \ll r_{\max}$ through contrastive pairings; (2) Amplification of policy update signals via pairwise comparisons rather than absolute reward thresholds; (3) Integration of temporally consistent state-action histories $\tau_{1:t-1}$ to maintain trajectory coherence. By learning from these constructed preference pairs, the policy $\pi_\theta$ develops enhanced discriminative capabilities for identifying and optimizing high-value trajectories, even under sparse environmental feedback conditions.

\subsection{Preference Fine-Tuning with CoT}
\label{sec:apw}

Our visual language model (VLM) processes embodied tasks through structured prompts containing environmental observations and action trajectories. As shown in Figure~\ref{fig:main_pic}, this framework enables contextual decision-making through chain-of-thought (CoT) reasoning followed by executable actions. The model's response sequence explicitly links cognitive processes with physical actions - each reasoning trajectory concludes with an \texttt{action} token denoting the selected operation, maintaining explicit action-rationale alignment.

Critical analysis reveals that conventional fine-tuning methods disproportionately prioritize action optimization while neglecting CoT coherence preservation. This imbalance disrupts the linguistic consistency established during pretraining, ultimately degrading reasoning capabilities. To address this dual challenge of action effectiveness and cognitive integrity, we implement two core constraints. First, we maintain distributional alignment between fine-tuned outputs and the reference model through KL-divergence control. Building on DPO principles, we establish:
\begin{equation}
\small
\label{dpo_q}
Q(s, a) = \beta \log{\frac{\pi_\theta(a|s)}{\pi_{ref}(a|s)}}
\end{equation}

Second, TCPO explicitly prioritizes reasoning quality over action selection during gradient updates. This approach simultaneously enhances decision reliability while preserving the model's inherent linguistic coherence - crucial for maintaining robust reasoning capabilities in dynamic environments. Then, we can fine-tune the output strategy of VLM by optimizing the $Q$ value, while limiting the output distance between the fine-tuned model and the reference model without fine-tuning by adding a regularization term of KL divergence to the optimization objective, which is as follows:
\begin{equation}
\small
\label{kl_op}
\max_{\pi_{\theta}} \mathbb{E}_{s\sim D, a\sim \pi_{\theta}(a|s)}[Q(s, a)] - \beta D_{\mathrm{KL}}[\pi_\theta \parallel \pi_{ref}] \\
\end{equation}

Based on this optimization objective, combined with some mathematical derivations of \cite{yang2024usinghumanfeedbackfinetune}, we can derive the following step-wise optimization formula:
\begin{equation}
\small
\label{naive_dpo}
\begin{aligned}
\mathcal{L} &= -\mathbb{E}_{\zeta} \log \sigma (  \beta \log \frac{p(a_1^t|\mathcal{T}_1^t) \pi_\theta(\mathcal{T}_1^t | \tau_1^{t-1})}{\pi_{ref}(a_1^t,\mathcal{T}_1^t | \tau_1^{t-1})} \\ 
&- \beta \log \frac{p(a_2^t|\mathcal{T}_2^t) \pi_\theta(\mathcal{T}_2^t | \tau_2^{t-1})}{\pi_{ref}(a_2^t,\mathcal{T}_2^t | \tau_2^{t-1})} )
\end{aligned}   
\end{equation}

\noindent where we have $\Lambda$ instead of $\sigma(\hat{Q}_\theta(a_1^t, \mathcal{T}_1^t, \tau_1^{t-1})-\hat{Q}_{\theta}(a_2^t, \mathcal{T}_2^t, \tau_2^{t-1}))$. The complete mathematical derivation is provided in Appendix~\ref{app:form-derivation}. Our formulation introduces $\mathcal{T}_i^t$ as the CoT reasoning text at step $t$, with $a$ denoting the resultant action. The gradient of the objective in Equation~\ref{naive_dpo} is:
\begin{equation}
\small
\begin{aligned}
\label{naive_dpo_grad}
\nabla_{\theta}\mathcal{L} &= -\beta\mathbb{E}_{\zeta} [\Lambda[ \nabla_{\theta}\log\pi_\theta(\mathcal{T}_1^t | \tau_1^{t-1}) \\ &- \nabla_{\theta}\log \pi_\theta(\mathcal{T}_2^t | \tau_2^{t-1})]]
\end{aligned}
\end{equation}
which reveals the elimination of direct action probability influences. This motivates our practical \textbf{TCPO} implementation with Action Probability Weighting (APW):
\begin{equation}
\small
\begin{aligned}
\tilde{\mathcal{L}} = &-\mathbb{E}_{\zeta} \log \sigma [ \beta p(a_1^t|\mathcal{T}_1^t) \log \frac{\pi_\theta(\mathcal{T}_1^t | \tau_1^{t-1})}{\pi_{ref}(a_1^t,\mathcal{T}_1^t | \tau_1^{t-1})} \\ &- \beta p(a_2^t|\mathcal{T}_2^t)\log \frac{ \pi_\theta(\mathcal{T}_2^t | \tau_2^{t-1})}{\pi_{ref}(a_2^t, \mathcal{T}_2^t | \tau_2^{t-1})} ]
\label{eq:aw-loss}
\end{aligned}
\end{equation}
The errors of both components and the feasibility of our approach will be analyzed later. 
Intuitively, the gradient term of the action probability serves to reinforce the probability of the corresponding thoughts. Actions with higher probabilities following CoT reasoning indicate stronger alignment with the underlying thought process, whereas lower probabilities suggest more randomness in action generation. This weighting mechanism helps suppress the generation of highly random positive samples and promotes the production of more deterministic, thought-aligned samples.

We provide a simple illustration of Equation \ref{eq:aw-loss} to demonstrate that the approximation is reasonable. Assuming that the pre-trained model has achieved good alignment, so $p(a|\mathcal{T})$ will be close to $1$. We have the following:

\begin{equation}
\small
\begin{aligned}
    &\Delta(p(a_i^t|\mathcal{T}_i^t)) = \log \frac{p(a_i^t|\mathcal{T}_i^t) \pi_\theta(\mathcal{T}_i^t|\tau_i^{t-1})}{\pi_{ref}(a_i^t, \mathcal{T}_i^t|\tau_i^{t-1})} \\
    &- p(a_i^t|\mathcal{T}_i^t)\log \frac{\pi_\theta(\mathcal{T}_i^t|\tau_i^{t-1})}{\pi_{ref}(a_i^t, \mathcal{T}_i^t|\tau_i^{t-1})} \\
    &=\log{p(a_i^t|\mathcal{T}_i^t)} + (1 - p(a_i^t|\mathcal{T}_i^t)) \log \frac{\pi_\theta(\mathcal{T}_i^t|\tau_i^{t-1})}{\pi_{ref}(a_i^t, \mathcal{T}_i^t|\tau_i^{t-1})}
\end{aligned}
\end{equation}
This variable will approach zero as $p(a|\mathcal{T})$ approaches $1$. In practice, we have calculated the approximate distribution of action probabilities and demonstrated that our assumption is well-founded, which is illustrated in Figure \ref{fig:aprob}.

\subsection{Action Policy Consistency}

In the second optimization stage, we introduce a regularization term to constrain the final action text output. Direct fine-tuning of the reasoning chain may inadvertently modify the model's inherent language generation patterns, potentially inducing catastrophic forgetting. By aligning the action text outputs with those of a reference foundation model, we ensure strict adherence to the prompt's structural requirements while maintaining action validity. This regularization mechanism preserves output integrity without compromising task performance, with empirical effects demonstrated in Figure~\ref{fig:kl-cases}. To strengthen the consistency between the Chain-of-Thought reasoning process and final action generation, we propose augmenting the optimization framework with an additional constraint. Our key insight stems from the observation that pre-trained language models already exhibit well-optimized mappings from reasoning traces to actions. During interactive learning phases, we therefore enforce alignment with these pre-trained behaviors through an \textbf{Action Policy Consistency (APC)} constraint, implemented via L2 regularization term:
\begin{equation}
\small
\label{eq:finloss}
    \mathcal{L}_{\text{TCPO}} = \tilde{\mathcal{L}} + \kappa \cdot \mathrm{L_2}\left(\pi_\theta(a_1^t|\mathcal{T}_1^t), \pi_{\text{ref}}(a_1^t|\mathcal{T}_1^t)\right)
\end{equation}
\noindent where $\kappa$ serves as a tunable hyperparameter controlling constraint intensity, and $\mathrm{L_2}(\cdot)\equiv\parallel\cdot\parallel_2$. Unlike the KL divergence typically used in DPO formulations, this constraint specifically targets the thought-to-action mapping process rather than overall output distribution matching.

To evaluate the effectiveness of the APC constraint, we perform comparative experiments examining model outputs at 2k training steps (Figure~\ref{fig:kl-cases}). The constrained model generates coherent CoT reasoning that logically leads to valid actions. In contrast, the unconstrained model often exhibits a disconnect between reasoning and action—despite producing plausible intermediate reasoning, it frequently results in invalid final actions unrelated to the preceding analysis. This divergence underscores the importance of enforcing explicit reasoning-action alignment to preserve decision consistency. The full training procedure for our method is provided in Appendix~\ref{Pseudo:Code:of:TCPO}.

\begin{table*}[t]
\centering
{\fontsize{8.4pt}{12pt}\selectfont
\begin{tabular}{
    >{\raggedright}p{3.6cm}
    |*{4}{c}  % gym_cards 列：EZP, P24, BJ, NL, Avg.
    |>{\columncolor[gray]{0.9}}c
    |*{6}{c}  % alfworld 列：Pick2, Look, Clean, Heat, Cool, Pick, Avg.
    |>{\columncolor[gray]{0.9}}c
    }
\toprule
\multirow{2}{*}{} & \multicolumn{5}{c|}{GymCards} & \multicolumn{7}{c}{ALFworld} \\
\cline{2-13}  % 调整合并范围
& EZP & P24 & BJ & NL & Avg. 
& Pick2 & Look & Clean & Heat & Cool & Pick & Avg. \\
\midrule
CNN+RL & 0 & 0 & 38.8 & \underline{87.1} & 31.5 & 0 & 0 & 0 & 0 & 0 & 0 & 0.0 \\
GPT4-V~\citep{yang2023dawnlmmspreliminaryexplorations} & 10.5 & 0 & 25.5 & 65.5 & 25.4 & 14.6 & 12.1 & \underline{18.8} & 6.7 & 17.8 & 38.2 & 19.4 \\
Gemini~\citep{geminiteam2024geminifamilyhighlycapable} & 2.0 & 0 & 30.0 & 82.5 & 28.6 & 12.0 & \underline{16.7} & 0 & 0 & 0 & 34.6 & 13.5 \\
LLaVA-sft~\citep{liu2024improvedbaselinesvisualinstruction} & 23.0 & 2.6 & 23.1 & 24.8 & 18.4 & \textbf{28.6} & 0 & 14.4 & 11.1 & 0 & \underline{39.2} & 17.7 \\
RL4VLM~\citep{zhai2024fine} & \underline{35.0} & \underline{7.0} & \underline{39.3} & \textbf{89.4} & \underline{42.7} & 20.6 & 15.1 & 10.0 & \underline{17.0} & 5.6 & 36.9 & \underline{20.0} \\
\midrule
\textbf{TCPO (Ours)} & \textbf{50.0} & \textbf{11.1} & \textbf{40.3} & 70.0 & \textbf{42.9} & \underline{27.3} & \textbf{33.3} & \textbf{25.0} & \textbf{28.6} & \underline{5.9} & \textbf{41.7} & \textbf{26.7} \\
\bottomrule
\end{tabular}
}
\caption{We present results demonstrating that fine-tuning the VLM using TCPO and PPO leads to varying task completion rates and average task completion rates in the GymCards tasks and ALFworld environment. Our findings show that, for most tasks, fine-tuning the model with preference-based methods outperforms reinforcement learning approaches in terms of task performance. Furthermore, we observe that the preference method achieves the same average task completion rate as PPO with fewer interaction steps, highlighting its higher sample efficiency and reduced model degradation during online interaction with the environment.}
\label{tab:main}
\vspace{-3mm}
\end{table*}

\section{Experiments}
\label{gen_inst}
To systematically evaluate our proposed framework, we design experiments addressing three core research questions:
\begin{itemize}[leftmargin=*]
    \item Does TCPO effectively enhance visual-language models' decision-making proficiency in embodied simulation environments?
    \item Can TCPO stabilize action distributions and prevent policy degradation in interactive learning?
    \item Does the action probability consistency constraint improve reasoning-action alignment in model outputs?
\end{itemize}

\paragraph{Experimental Setup} 
Our empirical evaluation utilizes the \texttt{gym\_cards} environment featuring four core tasks: \textit{Number Line} (\textbf{NL}), \textit{Easy Pick} (\textbf{EZP}), \textit{Pick-24} (\textbf{P24}), and \textit{Blackjack} (\textbf{BJ}) and \texttt{alfworld} benchmark environment \citep{shridhar2020alfred}, containing six distinct household task categories: \textit{Pick \& Place} (abbreviated as \textbf{Pick}), \textit{Pick Two \& Place} (\textbf{Pick2}), \textit{Clean \& Place} (\textbf{Clean}), \textit{Cool \& Place} (\textbf{Cool}), \textit{Heat \& Place} (\textbf{Heat}), and \textit{Examine in Light} (\textbf{Look}). Each task requires agents to process egocentric visual observations and textual instructions for sequential navigation and manipulation. The implementation builds upon the LLaVA-v1.6-Mistral-7B architecture \citep{liu2023visualinstructiontuning}, extended with our TCPO framework. Visual observations are processed through a structured input pipeline that serializes multi-modal inputs into model-compatible prompts while preserving original instruction-following capabilities.
Our analysis focuses on ALFWorld's enhanced complexity, where agents perform multi-step household operations requiring sequential manipulation and spatial reasoning. This environment is prioritized for its comprehensive benchmarking that better reflects real-world challenges compared to foundational GymCards tasks.

\begin{figure}[t]
    \centering
    \includegraphics[width=1\linewidth]{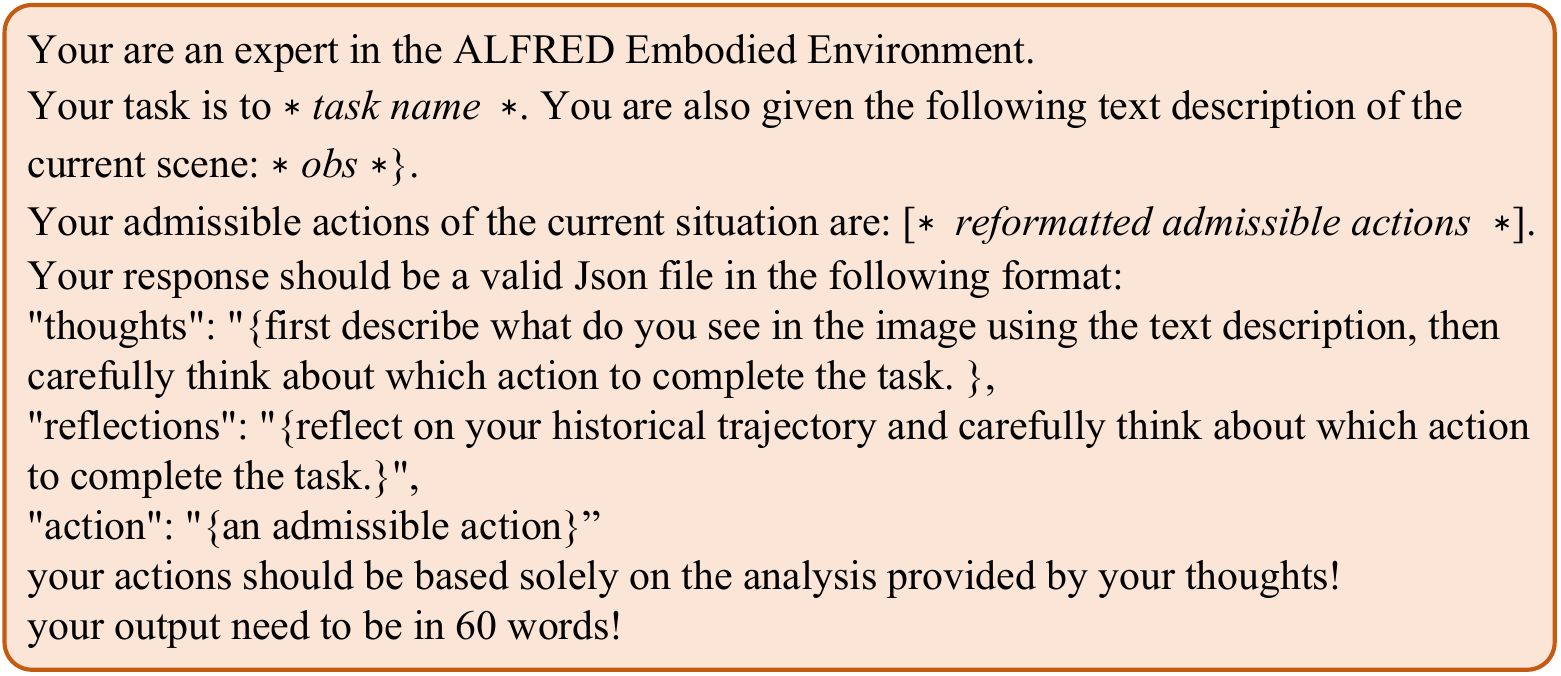}
    \caption{Prompt used in ALFWorld tasks.}
    \label{fig:pt}
    \vspace{-2mm}
\end{figure}

\paragraph{Prompt Design} 
Our chain-of-thought prompting strategy integrates three core components through natural language instructions. First, we formalize ALFWorld's semantic instructions into structured objectives, mapping paraphrased commands like "examine the pillow with the desklamp" and "look at the pillow under the desklamp" to standardized procedural sequences involving object localization, navigation, and interaction. Second, we explicitly define action space constraints based on environment dynamics, specifying preconditions (e.g., proximity requirements for object interaction) and postconditions (e.g., possession prerequisites for placement actions), with action validity forming a key evaluation metric. Finally, we enforce strict JSON output formatting requiring logically connected \texttt{thoughts} and \texttt{action} fields, ensuring causal relationships between reasoning traces and final decisions, while rigorously observing output text length constraints. The complete prompt structure with exemplar inputs is visualized in Figure~\ref{fig:pt}.

\paragraph{Implementation} 
Our training integrates supervised fine-tuning (SFT) on the LEVI-Project/sft-data corpus \citep{zhai2024fine} containing 45k GPT-4-generated expert trajectories, ensuring structured JSON outputs for action-thread consistency. Subsequent environmental interaction employs online policy optimization with real-time monitoring of action validity and trajectory coherence, dynamically adjusting learning parameters to preserve structured response patterns while enhancing decision-making in interactive scenarios.
\begin{figure}[t]
    \centering
    \includegraphics[width=0.9\linewidth]{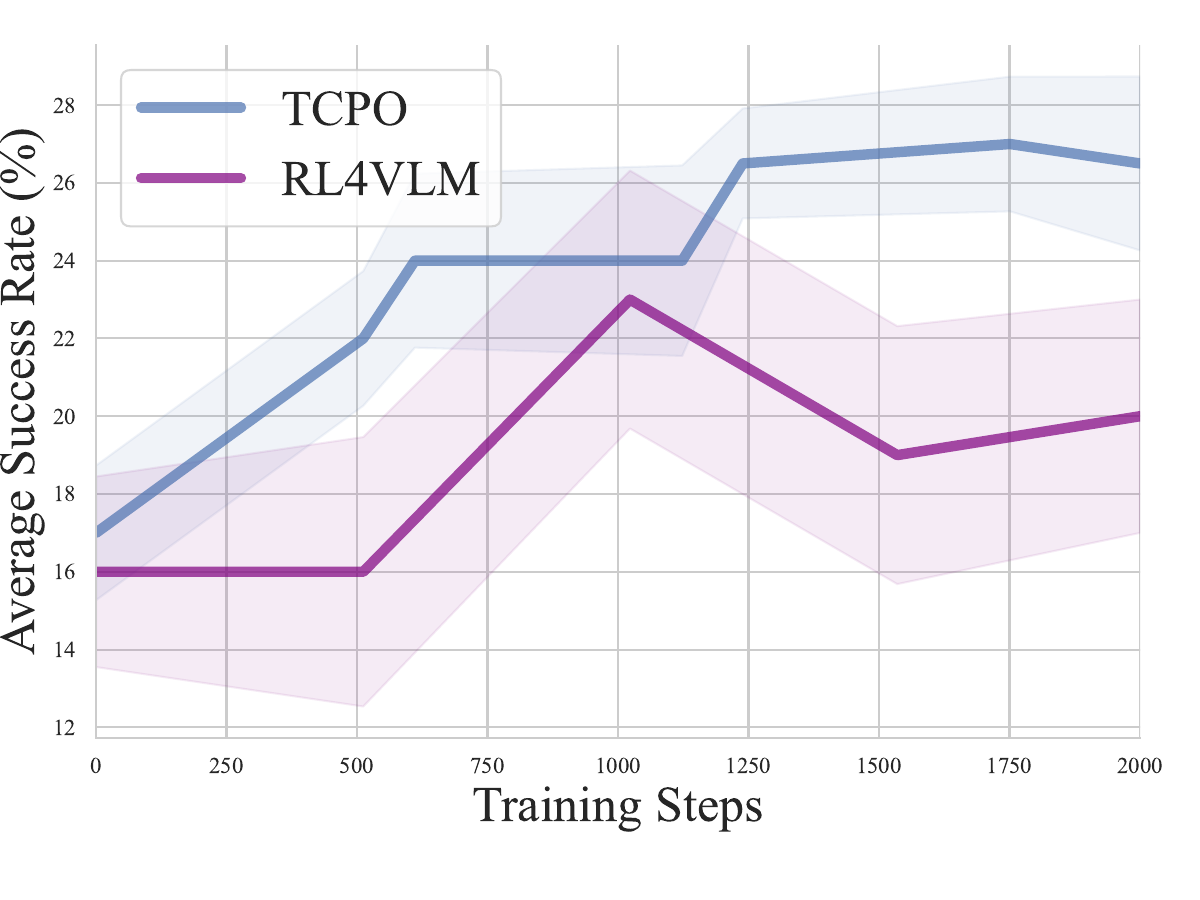}
    \vspace{-6mm}
    \caption{Training curves in the ALFWorld environment. In the first 2k steps, the TCPO method demonstrates superior convergence and efficiency compared to the PPO-based interaction method.}
    \label{fig:main_curves}
    \vspace{-4mm}
\end{figure}
\subsection{How much better we are at making decisions}
\label{exp1}

The aim of experiments in this section is to validate the performance of TCPO. We first conduct preliminary validation on GymCards' four core tasks (NL, EZP, P24, BJ), where TCPO achieves 42.9\% average success rate compared to 42.7\% of PPO, demonstrating superior instruction comprehension in constrained scenarios. To evaluate whether the algorithm can consistently generate decisions through the CoT process, we use the success rate of task execution as a reference and select PPO from the RL4VLM \citep{zhai2024fine} framework as the baseline. Our baseline results for CNN+RL, GPT4-V, Gemini, and LlaVA-sft are directly reused from RL4VLM due to the lack of reproduction details. In contrast, the RL4VLM baseline was rigorously reproduced using the original training methods and parameters. In GymCards experiments, we implement task-specific reward shaping where preference scores incorporate both game completion and strategic depth metrics, using Equation \ref{eq:preference} with adjusted weights for card-game dynamics. ALFWorld does not provide a reward function during interactions, it only indicates whether the current task is successfully executed and returns the task's progress. 
Given that such progress updates are sparse in a larger action space, we construct preference criteria for preference learning. The preference score for each trajectory is calculated using Equation \ref{eq:preference}:
\begin{equation}
\small
\label{eq:preference}
    P = 50 * success\ rate - \mathbb{1}_{\{invalid\}}
\end{equation}
\[
\mathbb{1}_{\{invalid\}} = 
\begin{cases} 
1 &  action\ \text{not}\ \text{in}\ admissible\ action \\ 
0 & \text{otherwise} 
\end{cases}
\]
\begin{table}[t]
  \caption{Performance comparison (average success rates \%) with preference-based learning approaches.}
  \label{tab:preference_comp}
  \centering
  \resizebox{0.9\linewidth}{!}{
  \begin{tabular}{l|cc}
    \toprule
     Methods    & GymCards          & ALFworld        \\
    \midrule
      PPO \cite{schulman2017proximalpolicyoptimizationalgorithms}    & 32.8  & 20.0 \\
      DPO \cite{rafailov2024directpreferenceoptimizationlanguage}    & 31.5  & 18.8  \\
      D3PO \cite{yang2024usinghumanfeedbackfinetune}    & 35.6  & 22.1 \\
      \midrule
      \textbf{TCPO (Ours)}   & \textbf{42.9}  & \textbf{26.7}  \\
    \bottomrule
  \end{tabular}
  }
  \vspace{-7mm}
\end{table}

\noindent where $\mathbb{1}_{\{invalid\}}$ represents the rejection of illegal actions given the same success rate. During the exploration phase, the agent collects trajectory data and constructs sample pairs based on the six task types mentioned above. Higher preference scores indicate greater sample preference. In practice, considering the achievement of long-term goals, we calculate preference scores using a method similar to discount factor weighting in reinforcement learning returns.
Due to the high randomness of ALFWorld, we set up experimental environments with different seeds and calculated the mean and variance of each result. 

We use Equation \ref{eq:finloss} for the model weight update with $\kappa=0.1$, measure the agent's performance by the average success rate of each task. The final comparisons are shown in Table \ref{tab:main} and \ref{tab:preference_comp}. It can be seen that TCPO consistently outperforms all RL-based and preference-based baselines.
We further plot the change in the average success rate over training on ALFWorld in Figure \ref{fig:main_curves}. TCPO exhibits a more robust growth, and the continuous rise of the curve confirms the improvement in model degradation issues.
\begin{figure*}[t]
    \centering 
    \subfloat[Impact of Kappa ($\kappa$)]{\includegraphics[width=0.7\columnwidth]{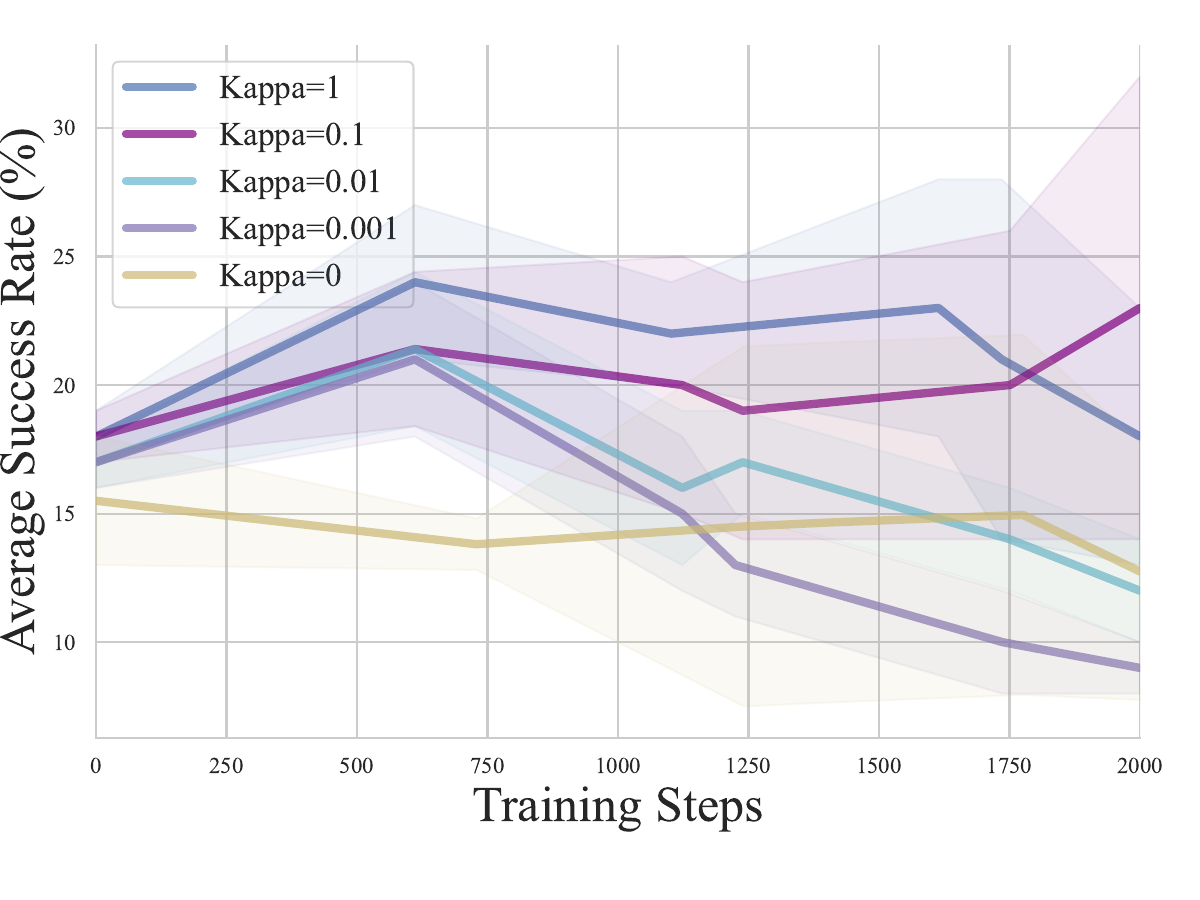}\label{fig:abl_kl}}
    \subfloat[Comparison of success rates]{\includegraphics[width=0.7\columnwidth]{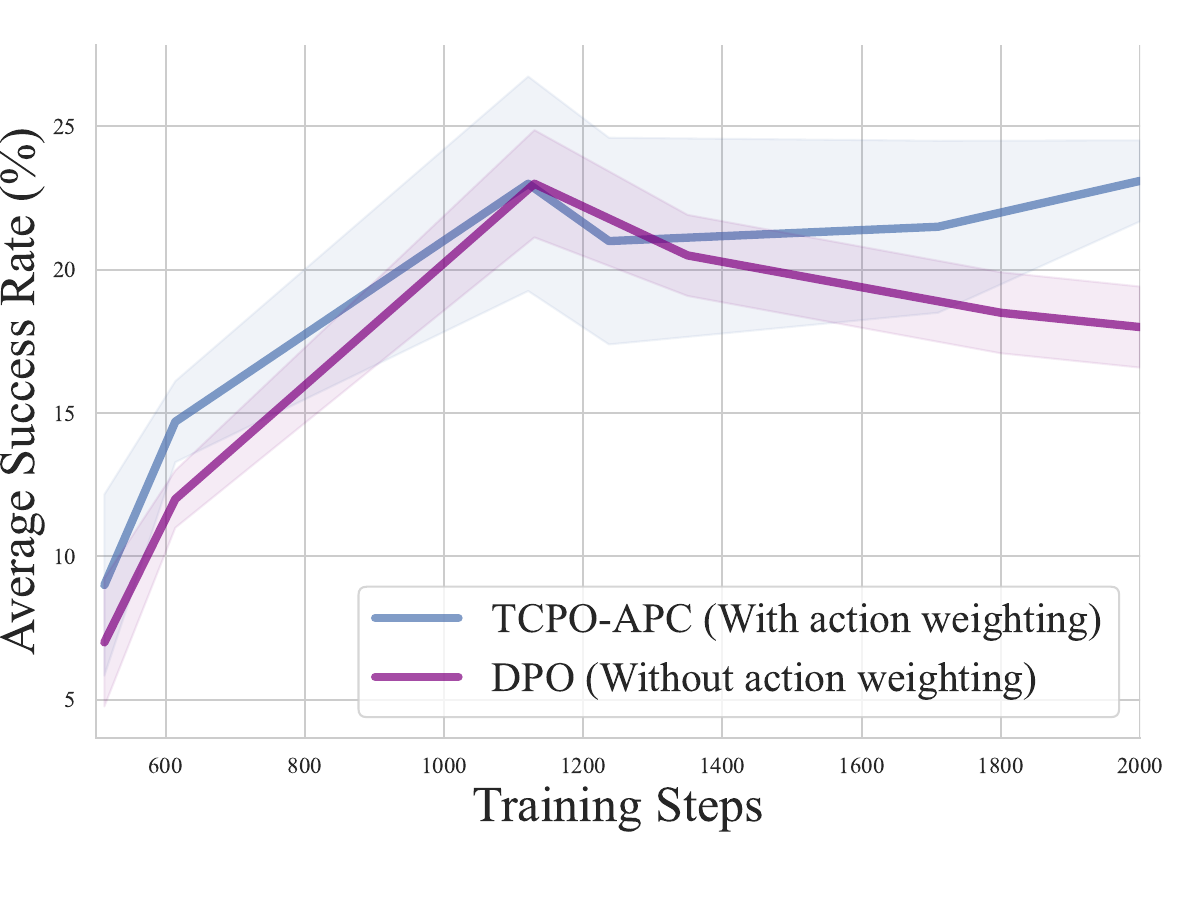}\label{fig:aw_curve}}
    \subfloat[Action token probability]{\includegraphics[width=0.68\columnwidth]{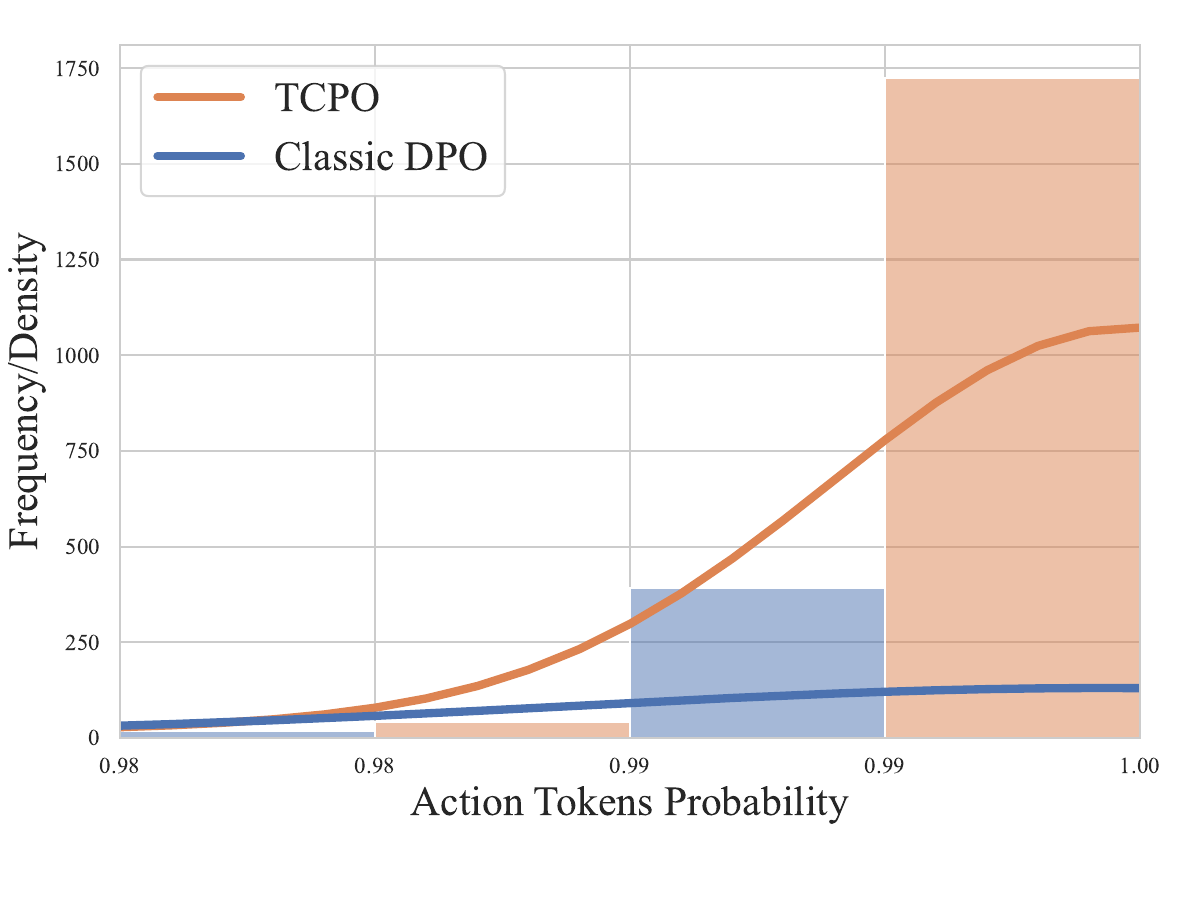}\label{fig:aprob}}
    \vspace{-1mm}
    \caption{The experimental result of TCPO. (a) Impact of different values of $\kappa$ in APC. (b) Comparison of the average success rates between TCPO-APC and classic DPO. (c) Action tokens probability distribution.}
    \label{fig:intro_case}
    \vspace{-3mm}
\end{figure*}
ALFWorld gives task randomly so we calculate the overall success rate as the weighted average of success rates under all tasks. TCPO shows an improvement in the overall success rate, indicating that our algorithm can learn more efficiently from interactions.
In our experiments, we used approximations such as $\log(\pi(a|\mathcal{T}, \tau)\pi(\mathcal{T}|\tau)) \approx \pi(a|\mathcal{T}, \tau)\log(\pi(\mathcal{T}|\tau))$ when $\pi(a|\mathcal{T}, \tau) \rightarrow 1$. We calculated the occurrence probability distribution of action tokens in the experiments to demonstrate that our approximations are reasonable.

\subsection{What role does action policy consistency constraint play?}
We pointed out that during training, to enhance stability, we introduced the regularization of action token probabilities between finetune model and reference model. This section will explore the impact of regularization on the results and investigate its role. We designed ablation experiments, where we conducted trials with different regularization weight values $\kappa$ under the same parameter settings, and recorded the average success rate of the agent during training. In this experiment, we use Equation \ref{naive_dpo} with the regular term as the loss function, with other conditions the same as in Section \ref{exp1}.

The results in Figure \ref{fig:abl_kl} show that different values of $\kappa$ significantly impact the success rate. As the parameter increases, the action policy consistency constraint strengthens, leading to improved model performance. This validates the importance of regularization. However, when $\kappa$ is set to 1, the algorithm's performance declines, indicating that $\kappa$ should neither be too large nor too small, with a value around 0.1 yielding near-optimal performance. Given the importance of the $\kappa$ parameter, its optimal value may vary across different environments or tasks. The optimal $\kappa$ value of 0.1 was determined through comprehensive testing across a three-order-of-magnitude parameter range. The experiment demonstrated exceptional robustness, maintaining consistent performance across diverse conditions without requiring scenario-specific parameter adjustments. Due to space constraints, we do not explore this further in this paper.

\subsection{APW in TCPO}

Our analysis in Section \ref{sec:apw} establishes that action probability weighting (APW) intrinsically reinforces decision determinism through cognitive-behavioral alignment. prioritizing gradient updates for high-probability actions. We validate this through comparative ablation studies between our APW-enhanced TCPO-APC framework (Equation~\ref{eq:aw-loss}) and baseline DPO (Equation~\ref{naive_dpo}), using identical experimental configurations. The comparative analysis of experimental results in Figure \ref{fig:aw_curve} delineates the performance comparison between the two experimental configurations, while Figure \ref{fig:aprob} quantitatively characterizes the action probability evolution during the initial 2000 training iterations. As demonstrated by the APW-conditioned results, the action token distribution exhibits predominant clustering near unity (probability $\approx$ 1), reflecting enhanced decision determinism and policy robustness. Conversely, the non-weighted configuration reveals a broadly distributed action probability spectrum, with notable instances of sub-0.9 probability values. Such dispersion in action selection probabilities suggests reduced policy convergence stability.
\begin{table}[t]
  \caption{Effect of joint optimization of TCPO.}
  \vspace{-2mm}
  \label{tab:optimization}
  \centering
  \resizebox{0.9\linewidth}{!}{
  \begin{tabular}{l|cc}
    \toprule
         & GymCards          & ALFworld        \\
    \midrule
      APW-only    & 35.8  & 23.0 \\
      APC-only   & 34.5  & 23.1  \\
      APC-APW-sequential    & 37.6  & 24.5 \\
      \midrule
      \textbf{TCPO}   & \textbf{42.9}  & \textbf{26.7}  \\
    \bottomrule
  \end{tabular}
  }
  \vspace{-5mm}
\end{table}
\subsection{Effect of Joint Optimization}
The overall objective of TCPO is a combination of APW and APC, which is jointly optimized to simultaneously enforce consistency between the Thoughts and Actions. To further validate the effectiveness of joint optimization, we present the experimental results in the Table \ref{tab:optimization}, comparing methods of APW only, APC only and sequential optimization of APW and APC. The results indeed demonstrate the effectiveness of the joint optimization of the Thoughts and Actions.

\section{Conclusions}
We introduce TCPO, an algorithmic framework for online interactive preference fine-tuning of multi-modal models during chain-of-thought reasoning. Built upon LLaVA-7B, TCPO achieves enhanced embodied task execution through dynamic replanning and rigorous CoT-action alignment via APW and APC. Experimental results demonstrate the superiority of TCPO over conventional reinforcement learning baselines in ALFWorld environments, with ablation studies confirming the critical role of APW in gradient prioritization and the contribution of APC to policy robustness.

% \newpage
\section*{Limitations}

Despite the effectiveness of our TCPO approach, there are two future directions that we'd like to point out. First, the Markovian assumption, as adopted in previous works, restricts the ability to handle complex non-Markovian decision processes in real-world scenarios. However, within our dynamically aligned environment algorithm, this assumption remains viable, as the algorithm inherently learns environmental dynamics. In this framework, redundant historical information may interfere with model judgment. Nevertheless, non-Markovian modeling is an important direction for future research. Moving forward, we plan to develop temporal modeling mechanisms to integrate historical information into TCPO, focusing on long-horizon task dependencies to eliminate the Markovian assumption. Second, empirical validation remains constrained to specific household tasks, necessitating broader domain evaluation to extend our approach to various domains. In the future, we plan to expand our experiments to a wider range of embodied environments such as VirtualHome.

\bibliography{latex/ref}

\clearpage
\appendix

\section{Training Details}
We provide more detailed hyper-parameters during in Table~\ref{tab:hparams}.
During SFT Phase, we use a dataset of 45k samples, with batch size 4 and 1 training epoch. The gradient accumulation steps is set to 1, with learning rate 2e-5. For the Online Learning Phase, the total sampling steps are 2k to 3k. Parameters are update 1 epoch training per online update cycle. 

Our baseline implementation strictly adheres to the reproduction methodology and parameters from the open-source RL4VLM project, as documented in their GitHub repository: \url{https://github.com/RL4VLM/RL4VLM}. The task completion rate is obtained through the data statistics of the interface feedback of the task completion rate in the environment. The environment determines task completion through its internal graph structure and returns a binary signal (0 for failure, 1 for success).
\begin{table}[h]
  \caption{Hyper-parameters for TCPO.}
  \label{tab:hparams}
  \centering
  \resizebox{0.9\linewidth}{!}{
  \begin{tabular}{cc}
\toprule
Hyperparameter & Value\\
\midrule
Seed	& 5 random seeds \\
Learning Rate	& 3e-4 \\
Mini Batch Size	& 1 \\
Grad Accum Steps& 	256\\
Max New Tokens &	1024\\
Temperature& 	0.2 \\
Discount Factor ($\gamma$)	& 0.99\\
Preference Weight ($\kappa$)&	0.1\\
Start Training Samples Nums	& 1,000\\
\bottomrule
  \end{tabular}
  }
\end{table}

\section{Derivation of Formulas}
\label{app:form-derivation}
We provide a simple derivation of Equation \ref{naive_dpo}.
During the RL phase with reward model, the object of training is to maximize returns. Following prior works the optimization is formulated as: 
\begin{equation}
    \max_{\pi_\theta} \mathbb{E}_{s\sim D, a\sim \pi_\theta(a|s)}[Q(s,a )] - \beta \mathrm{D_{KL}}[\pi_\theta \parallel \pi_{ref}]
\end{equation}
\noindent which can be rewritten as:
\begin{align*}
    \max_{\pi_\theta} &\mathbb{E}_{s\sim D, a\sim \pi_\theta(a|s)}[Q(s,a )] - \beta \mathrm{D_{KL}}[\pi_\theta \parallel \pi_{ref}] \\
    &= \max_{\pi_\theta} \mathbb{E}[Q(s,a)-\beta \log{\frac{\pi(a|s)}{\pi_{ref}(a|s)}}] \\
    &=\min_{\pi_\theta} \mathbb{E} [\log{\frac{\pi(a|s)}{\pi_{ref}(a|s)}} - \frac{1}{\beta}Q(s,a)] \\
    &=\min_{\pi_\theta} \mathbb{E} [\log{\frac{\pi(a|s)}{\pi_{ref}(a|s)\exp{(\frac{1}{\beta}Q(s,a))}}}] \\
    &= \min_{\pi_\theta} \mathbb{E}_{s\sim D}[\mathrm{D_{KL}}[\pi(a|s) \parallel \tilde{\pi}(a|s)]] 
\end{align*}
\noindent where $\tilde{\pi}(a|s)=\pi_{ref}(a|s)\exp{(\frac{1}{\beta}Q(s,a))}$. KL-divergence is minimized at zero if and only if the two distributions are identical. Therefore, in the case of the optimal solution we get:
\begin{equation*}
    \pi(a|s)=\tilde{\pi}(a|s)=\pi_{ref}(a|s)\exp{(\frac{1}{\beta}Q(s,a))}
\end{equation*}
A simple transformation yields: 
\begin{equation}
\label{eq:optmizeQ}
    Q(s,a)=\beta \log{\frac{\pi(a|s)}{\pi_{ref}(a|s)}}
\end{equation}
We can know from \citet{yang2024usinghumanfeedbackfinetune} that the Q-value form of Bradley-Terry preference distribution can be expressed as:
\begin{align}
\label{eq:BT}
    &p(\tau_1 > \tau_2 | a_i^{t}, s_i^{t}, a_i^{t-1}..., s_i^0)_{i \in \{{1,2}\}} \notag \\&= \frac{\exp(Q(s_1^t, a_1^t))}{\sum_{i \in \{{1, 2}\}}\exp{(Q(s_i^t, a_i^t))}}
\end{align}
Combining Eq. \ref{eq:optmizeQ} and Eq. \ref{eq:BT}, replacing $s_i^t$ with $\tau_i^{t-1}$ and $a_i^t$ with $(a_i^t, \mathcal{T}_i^t)$, we derive the following loss function:
\begin{align}
\mathcal{L} = -\mathbb{E}_{\zeta} \log \sigma   [\beta \log \frac{\pi_\theta(a_1^t, \mathcal{T}_1^t |\tau_1^{t-1})}{\pi_{ref}(a_1^t,\mathcal{T}_1^t | \tau_1^{t-1})} \notag \\ - \beta \log \frac{\pi_\theta(a_2^t, \mathcal{T}_2^t | \tau_2^{t-1})}{\pi_{ref}(a_2^t,\mathcal{T}_2^t | \tau_2^{t-1})}]
\end{align}
\noindent which is similar to Eq. \ref{naive_dpo}

\begin{table}[h]
  \caption{Impact of $\kappa$ on ALFWorld.}
  \label{tab:kappa}
  \centering
  \resizebox{1\linewidth}{!}{
  \begin{tabular}{ccccc}
\toprule
task &	$\kappa$ = 0.001 &	$\kappa$ =0.01 &	$\kappa$=0.1 &	$\kappa$=1 \\
\midrule
Pick &	25.7\%	&27.5\%&	41.7\%	&35.0\%\\
Pick2&	16.7\%	&25.0\%	&27.3\%	&18.5\%\\
Clean	&11.0\%	&22.2\%	&25.0\%	&10.0\%\\
Look	&25.0\%	&26.0\%	&33.3\%	&26.2\%\\
Heat	&6.8\%	&11.0\%	&28.6\%	&13.0\%\\
Cool&	1.0\%	&5.8\%	&5.9\%	&5.1\%\\
\midrule
Avg.	&13.3\%	&20.8\%	&26.7\%	&18.8\%\\
\bottomrule
  \end{tabular}
  }
\end{table}
\section{Parameter study of $\kappa$}
To further understand the impact of the parameter $\kappa$, we’ve conducted parameter experiments to observe the effects of varying $\kappa$ values under different tasks. The experimental results are shown in the Table \ref{tab:kappa}. It can be seen that $kappa$=0.1 achieves the best performance across all tasks in ALFWorld.

\section{Pseudo Code of TCPO}
\label{Pseudo:Code:of:TCPO}
We present the pseudo code of TCPO below for better understanding of our approach.
\begin{breakablealgorithm}
\caption{Training pipeline of embodied VLMs thorogh TCPO}
\begin{algorithmic}
    \REQUIRE Reference policy network $\pi_{ref}$, finetune policy network $\pi_{\theta}$, current observation $o_t$, past trajectories wises buffer $\tau_i^{t-1}$, $\{i=1\dots N\}$
    \FOR{$i = 1, 2, ..., N$}
    \STATE Randomly sample past trajectory $\tau_i^{t-1}$. The current chain of reasoning thoughts $\pi_{\theta}(\mathcal{T}_i^t | \tau_i^{t-1})$ and probabilities of output actions $p(a_i^t|\mathcal{T}_i^t)$ are generated through previous trajectories and fine-tuned models. $\pi_{ref}(a_i^t, \mathcal{T}_i^t | \tau_i^{t-1})$ is also obtained through the reference model.
    \FOR{$j$ in past trajectories buffer which can be paired with $\tau_i^{t-1}$.}
    \STATE The current chain of reasoning thoughts $\pi_{\theta}(\mathcal{T}_j^t | \tau_j^{t-1})$ and probabilities of output actions $p(a_j^t|\mathcal{T}_j^t)$ are generated through previous trajectories and fine-tuned models. $\pi_{ref}(a_j^t, \mathcal{T}_j^t | \tau_j^{t-1})$ is also obtained through the reference model.
    \STATE Calculating the loss in Equtation \ref{eq:aw-loss}.
    \ENDFOR
    \STATE Compute the Regularization loss $\mathrm{L_2}\left(\pi_\theta(a_1^t|\mathcal{T}_1^t), \pi_{\text{ref}}(a_1^t|\mathcal{T}_1^t)\right)$ and obtain the $\mathcal{L}_{\text{TCPO}}$.
    \ENDFOR
    \STATE The parameters $\theta$ are updated by backpropagation through the loss function.
\end{algorithmic}
\end{breakablealgorithm}

\begin{table}[h]
  \caption{Sample Efficiency Results.}
  \label{tab:sample_efficiency}
  \centering
  \resizebox{1\linewidth}{!}{
  \begin{tabular}{ccc}
\toprule
method &	success rate: 18\% &	success rate: 20\% \\
\midrule
PPO	& 650 (avg steps)	&810\\
DPO	&580	&670\\
TCPO&	400	&620\\
\bottomrule
  \end{tabular}
  }
\end{table}

\section{Effect of Sample Efficiency}
To better demonstrate the sample efficiency described in Table 1, we have conducted additional experiments to validate the sample. The training steps required by different methods to achieve varying average success rates during training are shown in Table \ref{tab:sample_efficiency}. As shown in the table, our algorithm achieves the same success rate while requiring fewer iteration steps through preference learning. 

\section{Design Details and Discussions}
\paragraph{Motivation of using $\mathrm{L_2}$ term in APC} We choose $\mathrm{L_2}$ loss for three primary reasons. First, computational efficiency - $\mathrm{L_2}$ term operates with O(n) computational complexity and offers simple implementation. Second, optimization stability - the linear gradients of $\mathrm{L_2}$ ensure higher stability in online algorithms and mini-batch optimization. Third, numerical robustness - $\mathrm{L_2}$ inherently avoids the need for log(0) protection mechanisms. In our experiments, we also tested KL divergence but observed inferior convergence performance compared to $\mathrm{L_2}$ term.

\paragraph{Sample pair construction scheme} Our preference sample pairs are generated through complete trajectory sampling. During experiments, we impose a trajectory length constraint by setting a maximum sampling step of 50. The preference score comprises two components: i) Final task success rate of the trajectory (0 or 1 in hard mode), and ii) Proportion of legal actions (trajectories with more legal actions are preferred under equivalent conditions). We estimate step-wise preference scores using a $\gamma$ weighting factor for credit assignment along the trajectory.

\paragraph{Success rate settings and discount factor weighting} The success rate signal is binary. The discount factor here serves for credit assignment. Since our preference construction is step-wise, it requires allocating contributions to each step within the same trajectory.

\paragraph{Train/test split} Our method employs online interaction for iterative updates, thus eliminating the need for train/test dataset split. During the online sampling process, we simultaneously calculate the agent's average success rate across all tasks and plot the corresponding success rate curve as shown in Figure 4. Specifically, our evaluation approach involves computing the average success rate once during subsequent sampling after each model update to assess current model capability. All tasks (the four Gymcards tasks individually, and all ALFWorld tasks collectively) include experimental data from at least 5 different seeds, with both mean values and variance displayed in the curves.
\onecolumn
\begin{table}
\caption{An example of the prompt and image in our tasks.}\label{fig:example}
\begin{tabular}{|p{0.99\textwidth}|}
\hline
\textbf{Inputs:} \\
You are an expert in the ALFRED Embodied Environment. You are also given the following text description of the current scene: `You arrive at loc 0. The cabinet 1 is open. On the cabinet 1, you see a pan 1, a kettle 1, a winebottle 1, a apple 1, a stoveknob 1, a stoveknob 2, a stoveknob 3, a stoveknob 4, a knife 1, a saltshaker 1, and a bread 1.`. Your task is to put a cool mug in cabinet. Your admissible actions of the current situation are: [`go to countertop 1', `go to cabinet 2', `go to countertop 2', `go to stoveburner 1', `go to drawer 1', `go to drawer 2', `go to drawer 3', `go to stoveburner 2', `go to stoveburner 3', `go to stoveburner 4', `go to drawer 4', `go to cabinet 3', `go to cabinet 4', `go to microwave 1', `go to cabinet 5', `go to cabinet 6', `go to cabinet 7', `go to sink 1', `go to sinkbasin 1', `go to fridge 1', `go to toaster 1', `go to coffeemachine 1', `go to cabinet 8', `go to drawer 5', `go to drawer 6', `go to drawer 7', `go to drawer 8', `go to shelf 1', `go to shelf 2', `go to countertop 3', `go to shelf 3', `go to drawer 9', `go to garbagecan 1', `open cabinet 1', `close cabinet 1', `take pan 1 from cabinet 1', `take kettle 1 from cabinet 1', `take winebottle 1 from cabinet 1', `take apple 1 from cabinet 1', `take stoveknob 1 from cabinet 1', `take stoveknob 2 from cabinet 1', `take stoveknob 3 from cabinet 1', `take stoveknob 4 from cabinet 1', `take knife 1 from cabinet 1', `take saltshaker 1 from cabinet 1', `take bread 1 from cabinet 1', `inventory', `look', `examine cabinet 1']. Your response should be a valid json file in the following format: \newline \{ \newline "thoughts": "first describe what do you see in the image using the text description, then carefully think about which action to complete the task.", \newline "action": "an admissible action" \newline \} \\
\hline
\textbf{Responses:} \\
\{ \\
"thoughts": "Based on the text description, I am at loc 0 with an open cabinet 1 in front of me. Items such as a pan, kettle, wine bottle, apple, several stoveknobs, a knife, saltshaker, and bread are visible on the cabinet 1. To find a cool mug to put in the cabinet, I may need to first locate a mug, which might be in one of the cabinets or drawers. Since the task does not involve any items currently in cabinet 1, and the mug is not mentioned in the description, my best course of action would be to start by searching the other cabinets or drawers for the mug.", \newline
"action": "go to cabinet 2" \\
\} \\
\hline
\end{tabular}
\end{table}
\twocolumn
\paragraph{Weighted average of success rates under all tasks} The ALFWorld environment contains 6 major task categories with over 5,000+ predefined task instructions, all of which are randomly assigned. Therefore, during sampling, we calculate the overall weighted average success rate by using the occurrence frequency of different task categories as weights. This approach helps reduce estimation bias - for instance, if tasks in the ``Pick" category were only executed once and succeed, their 100\% category success rate would significantly impact the arithmetic mean and increase the variance of estimated values.

\paragraph{Validity of the approximation in Equation \ref{eq:aw-loss}} In Figure 5(c), the X-axis and Y-axis represent Action Token Probability and Sample Density respectively, illustrating the probability distribution of action tokens across all sampled trajectories. Comparing TCPO and DPO, the TCPO method shows a probability distribution of final decision action tokens concentrated around 1, demonstrating that (a) the approximation condition in Section 3.2 is easily satisfied, and (b) TCPO naturally guides the model toward more deterministic action generation during training, supporting the reasonableness of the approximation.

\paragraph{Design of the prompt} The detailed description of the prompt used in our experiments is shown in Figure \ref{fig:pt}. Our approach intentionally requires the agent to verbalize its visual perceptions, fostering deeper situational awareness and more deliberate planning—significantly enhancing contextual comprehension. While direct planning without perceptual descriptions is technically feasible, this design choice strengthens reasoning. Additionally, we explicitly confirm that our framework does not incorporate any environmental descriptions beyond the agent’s own perceptual outputs. A example is shown in Table \ref{fig:example}.

\end{document}